\documentclass{article}

     \usepackage[preprint]{neurips_2021}

\usepackage[utf8]{inputenc} % allow utf-8 input
\usepackage[T1]{fontenc}    % use 8-bit T1 fonts
\usepackage{hyperref}       % hyperlinks
\usepackage{url}            % simple URL typesetting
\usepackage{booktabs}       % professional-quality tables
\usepackage{amsfonts}       % blackboard math symbols
\usepackage{nicefrac}       % compact symbols for 1/2, etc.
\usepackage{microtype}      % microtypography
\usepackage[table]{xcolor}         % colors
\usepackage{svg, graphics}
\usepackage{amsmath,graphicx}
\usepackage{enumerate}
\usepackage{diagbox}
\usepackage{enumitem}
\usepackage{multirow}
\usepackage{comment}
\usepackage{lipsum}
\usepackage{subcaption}
\usepackage[super]{nth}
\usepackage[ruled,vlined]{algorithm2e}
\usepackage{graphicx}
\usepackage{wrapfig}

\title{SpeechNet: a universal modularized model for speech processing tasks}

\author{Yi-Chen Chen$^{1,2}$, Po-Han Chi$^{1}$\thanks{* contribute equally}\ , Shu-wen Yang$^{1*}$, Kai-Wei Chang$^{1*}$, Jheng-hao Lin$^{1*}$, \\ \textbf{Sung-Feng Huang$^1$, Da-Rong Liu$^1$, Chi-Liang Liu$^1$, Cheng-Kuang Lee$^2$, Hung-yi Lee$^1$} \\
  $^1$National Taiwan University, Taiwan \ \ \ $^2$NVIDIA AI Technology Center, NVIDIA \\
  \texttt{$^1$\{f06942069,r08942074,r08944041,r09921048,r08922049,f06942045,} \\ \texttt{f07942148,r07942083,hungyilee\}@ntu.edu.tw} \ \ \texttt{$^2$ckl@nvidia.com} \\}

\begin{document}

\maketitle

\begin{abstract}
There is a wide variety of speech processing tasks ranging from extracting content information from speech signals to generating speech signals. 
For different tasks, model networks are usually designed and tuned separately. 
If a universal model can perform multiple speech processing tasks, some tasks might be improved with the related abilities learned from other tasks. 
The multi-task learning of a wide variety of speech processing tasks with a universal model has not been studied. 
This paper proposes a universal modularized model, SpeechNet, which treats all speech processing tasks into a speech/text input and speech/text output format. We select five essential speech processing tasks for multi-task learning experiments with SpeechNet. 
We show that SpeechNet learns all of the above tasks, and we further analyze which tasks can be improved by other tasks. 
SpeechNet is modularized and flexible for incorporating more modules, tasks, or training approaches in the future. 
We release the code and experimental settings to facilitate the research of modularized universal models and multi-task learning of speech processing tasks.
\end{abstract}

\section{Introduction}
\label{sec:intro}

%There is a wide variety of speech processing tasks, for example, automatic speech recognition (ASR), speech enhancement (SE), speaker classification (SC), text-to-speech (TTS) synthesis,  and voice conversion (VC), etc. For different tasks, model networks are usually designed and tuned separately, each aiming to optimize a specific metric. However, when we only focus on one task, we ignore some useful information provided by related tasks that might help the model do better on the original metric.

There is a wide variety of speech processing tasks, for example, automatic speech recognition (ASR), speech enhancement (SE), speaker classification (SC), text-to-speech (TTS) synthesis,  and voice conversion (VC), etc. 
These tasks involve different capabilities related to speech processing ranging from extracting content information from speech signals to generating speech signals.  
In literature, model networks are usually designed and tuned separately for different tasks, and each aims to expert a specific ability for processing speech. 
However, when we only focus on one task, we may ignore some useful abilities that can be shared across tasks to make the tasks better.
Human can learn different speech tasks and transfer the knowledge of different abilities between tasks. 
Can we train a universal model that can learn all the different speech processing abilities jointly in one model?

In this paper, we propose SpeechNet, a universal model for various speech processing tasks. 
Inspired by T5~\citep{raffel2019exploring}, we treat all speech processing tasks as the format: a task that takes speech/text input and produces speech/text output.
In SpeechNet, there are basic modules to handle different modalities, as illustrated in Figure~\ref{fig:speechnet} and introduced below:  
\begin{itemize}
\item Speech input: We use Prosody Encoder, Speaker Encoder and Content Encoder to extract prosody, speaker and content embeddings from speech.
\item Speech output: We use Audio Decoder to synthesize audio.
\item Text input: We use Text Encoder to map the input text to the content embedding space (which is the same output space of Content Encoder).
\item Text output: We use Text Decoder to produce text according to content embedding.
\end{itemize}
Most of the speech processing tasks can be done by concatenating the modules above, making multi-task learning (MTL) for a wide variety of speech processing tasks possible. It has been shown that the universal models trained to solve multiple tasks can benefit from multi-task learning (MTL)~\citep{vandenhende2020multi,zhang2017survey,ruder2017overview}, improving the generalizability and performance of models in NLP and computer vision.
During MTL, the tasks share the same modules, and the gradients computed from different objective functions of these tasks are accumulated to update the shared modules. 

This paper shows that SpeechNet can simultaneously learn five common and important speech processing tasks: ASR, SE, SC, TTS, and VC.
We conduct experiments with commonly used datasets for the five tasks. 
Based on the results of MTL, we know which combinations of speech tasks are effective.
SpeechNet is modularized and flexible for incorporating more modules, tasks, and training criteria in the future.  
We release the code\footnote{https://github.com/grtzsohalf/SpeechNet-codebase} and experimental settings to facilitate the research of universal modularized models or MTL of speech processing tasks.

\section{Related Work}
%Human brain is a system that can handle many tasks with different modalities.
It has been shown that a single deep learning model can jointly learn a number of large-scale tasks from multiple domains~\citep{kaiser2017one}.
\citep{mccann2018natural} proposes a model to solve ten tasks in natural language processing (NLP).
The core idea of the T5 model~\citep{raffel2019exploring}, a unified framework for a variety of text-based language problems, is to treat every text processing problem as a "text-to-text" problem, i.e., taking text as input and producing new text as output.

In the speech domain, some previous works train a model to solve two tasks or use an auxiliary task to improve the primary task's performance.
\citep{chen2015speech} studies SE and ASR.
\citep{tjandra2017listening,ren2019almost} study the duality of ASR and TTS. 
Some papers~\citep{jain2018improved,chen2020aipnet} use SC to improve ASR. \citep{tang2016multi} studies ASR and SC. \citep{hsu2019disentangling,jia2018transfer} shows the effectiveness of SC to help TTS. \citep{zhang2019improving} shows the performance of VC can be improved with text supervision.
\citep{kinnunen2017non} connects SC and VC. \citep{zhang2019joint} jointly trains TTS and VC. \citep{zhang2019non} uses TTS and SC to improve VC. 

%However, non of the above papers include more than two tasks.
However, in these works, models are designed for some specific speech processing tasks and not applicable for more speech processing tasks.
Moreover, the rapid rate of progress and diversity of techniques also make it difficult to compare different algorithms, tease apart the effects of new contributions, and understand the effectiveness of learning multiple speech processing tasks with one model.

\section{SpeechNet: a universal modularized model for speech processing tasks}
\label{sec:speechnet}

\begin{figure*}[tb!]
  \centering
  \centerline{\includegraphics[width=0.95\linewidth]{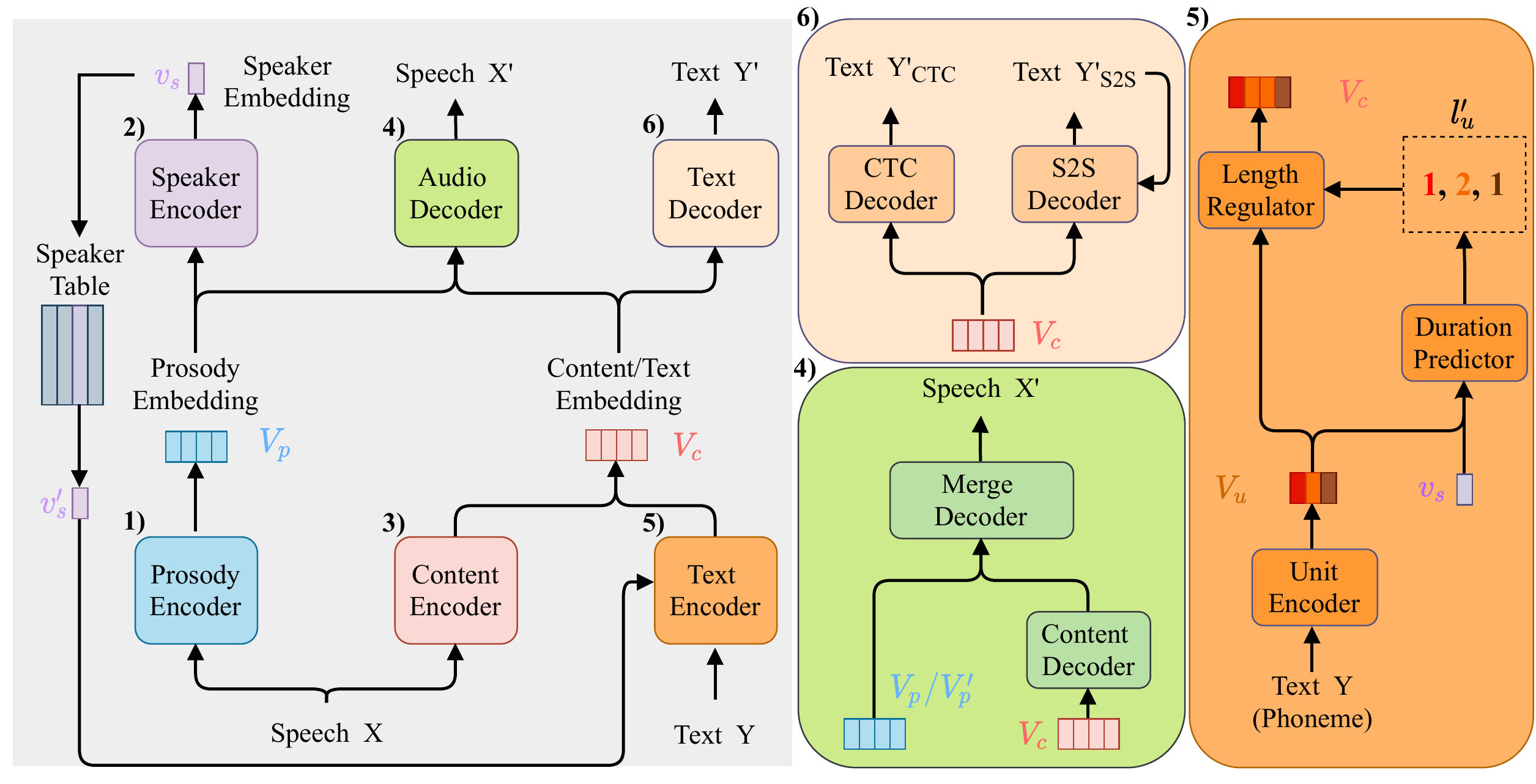}}
\caption{The architecture of SpeechNet is on the left. The six core modules in SpeechNet are Prosody Encoder, Speaker Encoder, Content Encoder, Audio Decoder, Text Encoder, and Text Decoder. More details about Audio Decoder, Text Encoder and Text Decoder are shown in a corresponding block on the right, with the same index and the same background color as that on the left of the figure. }
\label{fig:speechnet}
\end{figure*}

Any speech processing task can be treated as taking speech or text as input or output. SpeechNet contains six core modules for speech and text, respectively, to handle different modalities. These six modules are introduced in Subsection~\ref{subsec:modules}. How to concatenate the modules to do the five tasks used in this paper is described in Subsection~\ref{subsec:tasks}. 
At the end of this section, we discuss two problems of TTS and VC in this framework and further propose a modified version of SpeechNet by adding one additional module, Prosody Predictor, in Subsection~\ref{subsec:speechnet++}.

\subsection{Modules in SpeechNet}
\label{subsec:modules}

Here we discuss the detailed formulation of each module presented in Figure~\ref{fig:speechnet} respectively, while describing the model architecture details in Subsection~\ref{subsec:architecture}.

\subsubsection{Prosody Encoder: $E_P$}
\label{subsubsec: EP}

When speech $\mathbf{X}=\{\mathbf{x_1}, ..., \mathbf{x_T}\}$ with frame length $T$ serves as the input, it can be passed through Prosody Encoder $E_P$ to obtain a frame-level prosody embedding vector sequence
\begin{align}
    \mathbf{V_p} = E_P(\mathbf{X}), 
    \label{eq:EP}
\end{align}
Here we want to encode all the speaker and prosody characteristics in speech into $\mathbf{V_p}$.

\subsubsection{Speaker Encoder: $E_S$}
\label{subsubsec: ES}

The prosody embedding vectors can be further passed through Speaker Encoder $E_S$ to obtain an utterance-level speaker embedding vector
\begin{align}
    \mathbf{v_s} = E_S(\mathbf{V_p}), 
    \label{eq:ES}
\end{align}
which represents the speaker characteristics in speech. 
%The speaker embedding technique is commonly used in tasks such as speaker recognition/verification (SV)~\citep{variani2014deep,snyder2018x}, multi-speaker TTS~\citep{gibiansky2017deep} or VC~\citep{qian2019autovc}. 

\subsubsection{Content Encoder: $E_C$}
\label{subsubsec: EC}

The speech input can also be passed through Content Encoder $E_C$ to get a sequence of frame-level content embedding vectors
\begin{align}
    \mathbf{V_c} = \{\mathbf{v_{c_1}}, ..., \mathbf{v_{c_{T'}}}\} = E_C(\mathbf{X}), 
    \label{eq:EC}
\end{align}
which contain content information in speech. $T'$ is the length of content embedding vectors, which can be equal to the original frame length of speech $T$ or be shorter for more compact embeddings.

\subsubsection{Audio Decoder: $D_A$}
\label{subsubsec: DA}

When the output is speech, Audio Decoder $D_A$ takes a sequence of prosody embeddings $\mathbf{V_p}$ and a sequence of content embeddings $\mathbf{V_c}$ as input and output the desired speech. Specifically, the content embedding is firstly transformed by Content Decoder $D_C$, concatenated with the prosody embeddings, and then passed into Merge Decoder $D_M$ to output the speech 
\begin{align}
    \mathbf{X'} &= D_A(\mathbf{V_p}, \mathbf{V_c}) = D_M([\mathbf{V_p}; D_C(\mathbf{V_c})]).
    \label{eq:DA}
\end{align}

\subsubsection{Text Encoder: $E_T$}
\label{subsubsec: ET}

When text $\mathbf{Y}=\{y_1, ..., y_L\}$ with length $L$ serves as input, it is firstly encoded into unit token vectors
\begin{align}
    \mathbf{V_u} = \{\mathbf{v_{u_1}}, ..., \mathbf{v_{u_L}}\} = E_U(\mathbf{Y}),
    \label{eq:EU}
\end{align}
through Unit Encoder $E_U$. 

The number of input text tokens is usually much smaller than the frame length of output speech. To match the content embedding encoded from text and speech, we need to predict the frame length of each unit token, and replicate the unit token vectors accordingly to obtain the frame-level content embedding vectors with the same length from speech. It is called the length regulation technique originally proposed in TTS~\citep{ren2019fastspeech}. Specifically, we use Duration Predictor $DP$ to predict the frame lengths of each text units according to the speaker characteristics
\begin{align}
    \mathbf{l'_u} &= \{l'_{u_1}, ..., l'_{u_L}\} = DP([\mathbf{V_u}; \mathbf{v_s}]),
    \label{eq:DP}
\end{align}
and replicates unit token vectors according to frame lengths through Length Regulator $LR$ to obtain frame-level content embedding vectors
\begin{align}
    \mathbf{V_c} &= \{\mathbf{v_{c_1}}, ..., \mathbf{v_{c_T}}\} = LR(\mathbf{V_u}, \mathbf{l'_u}), \text{where}\ T = \Sigma_{i=1}^L l'_{u_i}.
    \label{eq:LR}
\end{align}
For example, if the text input sequence and corresponding unit token vectors are $\{a, b, c\}$, $\{\mathbf{v_{a}}, \mathbf{v_{b}}, \mathbf{v_{c}}\}$ respectively and the predicted frame lengths $\{1, 2, 1\}$, the content embedding vectors are $\mathbf{V_c} = \{\mathbf{v_a}, \mathbf{v_b}, \mathbf{v_b}, \mathbf{v_c}\}$.

It is worth noting that although the speaker embedding vector $\mathbf{v_s}$ is used during the generation of content embedding vectors, it is only used for duration prediction according to speaker characteristics and replication of text vectors. Therefore, each content embedding vector does not contain speaker information.
Overall, the generation of content embedding vectors through Text Encoder $E_T$ can be described as
\begin{align}
    \mathbf{V_c} &= E_T(\mathbf{Y}, \mathbf{v_s}) = LR(E_U(\mathbf{Y}), DP([E_U(\mathbf{Y}); \mathbf{v_s}])).
    \label{eq:ET}
\end{align}

\subsubsection{Text Decoder: $D_T$}
\label{subsubsec: DT}

When the output is text, Text Decoder $D_T$ takes the content embedding vectors as input and output the text sequence. In recent state-of-the-art sequence-to-sequence ASR models, two text sequences are decoded as output by two decoders: sequence-to-sequence (S2S) and connectionist temporal classification (CTC) decoders $D_{S2S}$ and $D_{CTC}$.
During the inference, we can simply select the text with the better decoder during training.
\begin{align}
    [\mathbf{Y'_{CTC}}; \mathbf{Y'_{S2S}}] &= D_T(\mathbf{V_c}) = [D_{CTC}(\mathbf{V_c}); D_{S2S}(\mathbf{V_c})].
    \label{eq:Y'}
\end{align}

\subsection{Five tasks for MTL}
\label{subsec:tasks}

\begin{figure*}[tb!]
  \centering
  \centerline{\includegraphics[width=\linewidth]{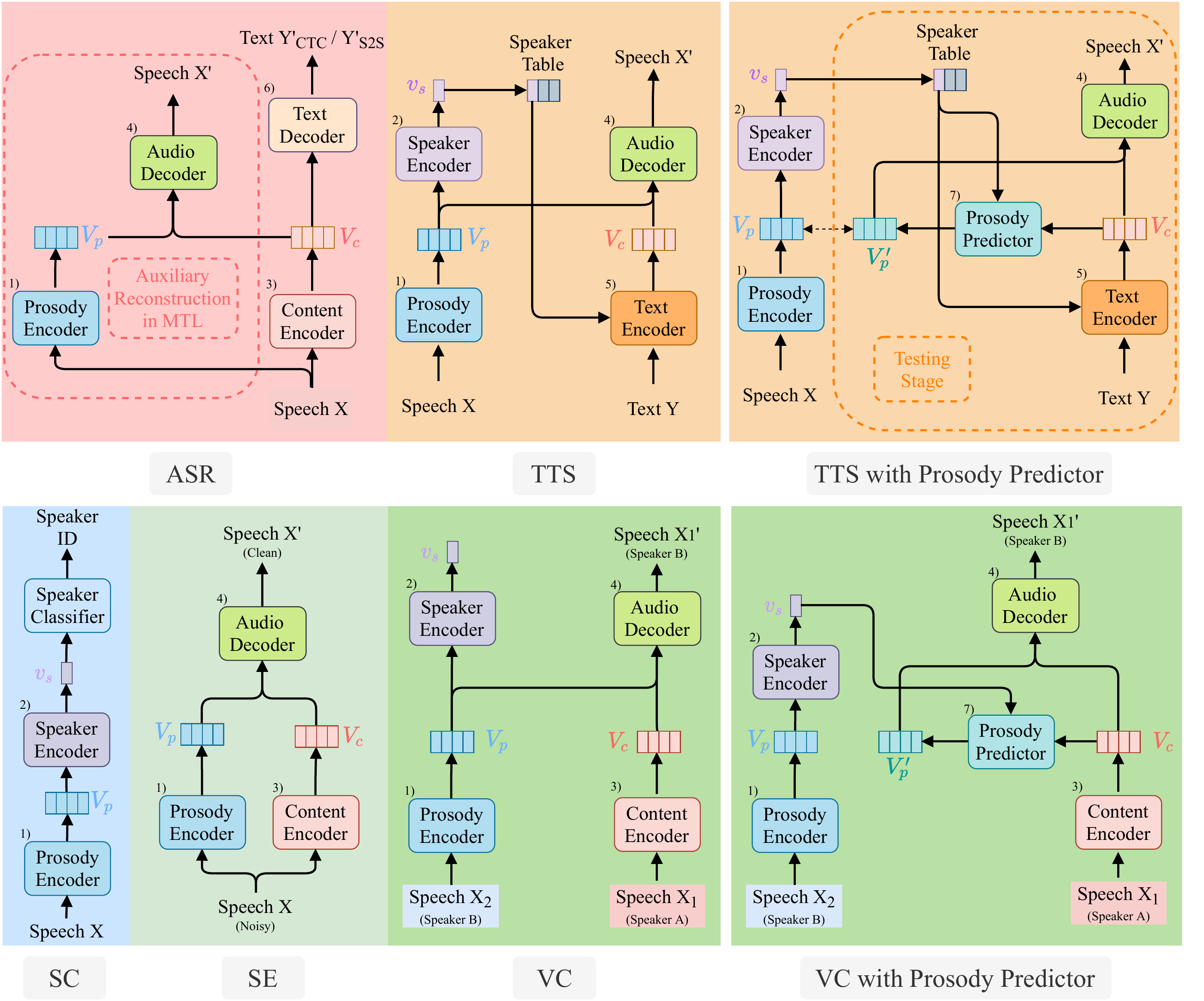}}
\caption{This figure shows how to combine the modules in SpeechNet into five different speech tasks. Each module block in this figure shares the same color and index in Figure~\ref{fig:speechnet}.}
\label{fig:tasks}
\end{figure*}

In this section, we describe how to combine the modules in SpeechNet into different speech tasks, which are also depicted in Figure~\ref{fig:tasks}.

\subsubsection{Automatic Speech Recognition (ASR)}
\label{subsubsec:asr}

In ASR, the input is speech $\mathbf{X}$ and the output is the corresponding transcription text $\mathbf{Y'_{CTC}}$ and $\mathbf{Y'_{S2S}}$:
\begin{align}
    [\mathbf{Y'_{CTC}}; \mathbf{Y'_{S2S}}] &= D_T(E_C(\mathbf{X})).
    \label{eq:asr}
\end{align}
The objective function is similar to those used in previous works~\citep{karita2019comparative,liu2019adversarial}, which is a weighted sum of a S2S loss and a CTC loss:
\begin{align}
    L_{ASR} = -\alpha_{ASR} \log P_{CTC}(\mathbf{Y'_{CTC}}|\mathbf{X}) - (1-\alpha_{ASR}) \log P_{S2S}(\mathbf{Y'_{S2S}}|\mathbf{X}),
    \label{eq:loss_asr}
\end{align}
where $P_{S2S}$ and $P_{CTC}$ are the S2S and CTC frame-wise posterior distributions of $\mathbf{Y'_{S2S}}$ and $\mathbf{Y'_{CTC}}$ given corresponding source $\mathbf{X}$ respectively, and $\alpha_{ASR}$ is a scalar hyperparameter.

In MTL, an auxiliary reconstruction objective function is also applied for aligning content vector space with other tasks:
\begin{align}
    L_{recon} =\ \parallel D_A(E_P(\mathbf{X}), E_C(\mathbf{X})) - \mathbf{X}\parallel^2
    \label{eq:loss_recon}.
\end{align}
The final loss for ASR is
\begin{align}
    L_{ASR\_total} = L_{ASR} + L_{recon}.
    \label{eq:total_loss_asr}
\end{align}

\subsubsection{Speech Enhancement (SE)}
\label{subsubsec:se}
In SE, the model takes noisy speech as input and outputs clean speech.
Here we encode the input noisy speech $\mathbf{X_{noisy}}$ into two parts, prosody embeddings and content embeddings, and decode back the denoised speech $\mathbf{X'}$:
\begin{align}
    \mathbf{X'} &= D_A(E_P(\mathbf{X_{noisy}}), E_C(\mathbf{X_{noisy}})).
    \label{eq:se}
\end{align}
The objective function is the mean absolute error (MAE) between the predicted and clean speech, $\mathbf{X'}$ and $\mathbf{X_{clean}}$, for more sensitivity to noise than mean square error (MSE):
\begin{align}
    L_{SE} = |\mathbf{X}' - \mathbf{X_{clean}}|.
    \label{eq:loss_se}
\end{align}

\subsubsection{Speaker Classification (SC)}
\label{subsubsec:sc}
In SC, the model takes speech as input and outputs the speaker identity.
Here we encode the input speech $\mathbf{X}$ into a speaker embedding vector, and use a speaker classifier $C_S$ to recognize the speaker $S'$:
\begin{align}
    S' &= C_S(E_S(E_P(\mathbf{X}))).
    \label{eq:sc}
\end{align}
The objective function is the cross entropy loss with regard to speaker labels:
\begin{align}
    L_{SC} = -\log P(S'|\mathbf{X}).
    \label{eq:loss_sc}
\end{align}

\subsubsection{Text-to-speech Synthesis (TTS)}
\label{subsubsec:tts}

%We perform speaker-adaptive TTS~\citep{fan2015multi}, where the model synthesizes speech conditioned on speaker characteristics. 
In TTS, we want to output a speech $\mathbf{X'}$ according to a text sequence $\mathbf{Y}$ conditioned on speaker characteristics.
Specifically, we maintain a trainable speaker embedding table, where every speaker in the training data corresponds to exactly one embedding vector in the table.
During training, we hope the speaker embedding output by Speaker Encoder to be as close as possible to the embedding in the table.
Therefore we have a MSE speaker loss:
\begin{align}
    L_{speaker} &= \parallel E_S(E_P(\mathbf{X})) - \mathbf{v'_s} \parallel,
    \label{eq:loss_speaker}
\end{align}
where $\mathbf{v'_s}$ is the corresponding speaker embedding of $\mathbf{X}$ in the table.

Then the overall TTS process becomes:
\begin{align}
    \mathbf{X'} &= D_A(E_P(\mathbf{X}), E_T(\mathbf{Y}, \mathbf{v'_s})).
    \label{eq:tts}
\end{align}
The objective function in training is the sum of (a) the mean square error (MSE) between the predicted and target speech $\mathbf{X'}$ and $\mathbf{X}$, (b) the MAE between the logarithms of predicted and original frame durations of text units, $\log(\mathbf{l'_u})$ and $\log(\mathbf{l_u})$ and (c) $L_{speaker}$:
\begin{align}
    L_{TTS} = \parallel\mathbf{X}' - \mathbf{X}\parallel^2 + |\log(\mathbf{l'_u}) - \log(\mathbf{l_u})| + L_{speaker}.
    \label{eq:loss_tts}
\end{align}
There are two problems with this setting. Firstly, during training, there is no additional constraint, so Prosody Encoder and Audio Decoder alone may become an autoencoder, and the content embeddings can be ignored.
Secondly, during inference, since the target speech is not available as input of Prosody Encoder, the input speech $\mathbf{X}$ has to be any other speech sentence uttered by the same speaker. However, the prosody of a speech is closely related to the content and duration of the speech.
Because the prosodies of input speech of Prosody Encoder and target speech do not match, the generated speech cannot be produced well. In Subsection~\ref{subsec:speechnet++}, we will address these two issues.

\subsubsection{Voice Conversion (VC)}
\label{subsubsec:vc}

In VC, we try to convert the voice of an audio clip from one speaker to another while preserving the content. 
Specifically, we input two speech utterances $\mathbf{X_1}$ and $\mathbf{X_2}$ with the same content but different speakers, and output the converted speech utterance $\mathbf{X'_{12}}$ with the content of $\mathbf{X_1}$ and speaker characteristics of $\mathbf{X_2}$
\begin{align}
    \mathbf{X'_{12}} &= D_A(E_P(\mathbf{X_2}), E_C(\mathbf{X_1})).
    \label{eq:vc}
\end{align}
Besides, to make the training more stable and easier, we also train the network to reconstruct the original utterances $\mathbf{X_1}$ and $\mathbf{X_2}$:
\begin{align}
    \mathbf{X'_1} &= D_A(E_P(\mathbf{X_1}), E_C(\mathbf{X_1})). \\
    \mathbf{X'_2} &= D_A(E_P(\mathbf{X_2}), E_C(\mathbf{X_2})).
    \label{eq:reconstruct}
\end{align}
The objective function is the sum of MSE losses of conversion and reconstruction:
\begin{align}
    L_{VC} &= \parallel \mathbf{X'_{12}} - \mathbf{X_{2}} \parallel^2 + \parallel \mathbf{X'_1} - \mathbf{X_1} \parallel^2 + \parallel \mathbf{X'_2} - \mathbf{X_2} \parallel^2.
    \label{eq:loss_vc}
\end{align}
During inference of this setting, since the target converted speech is not available as input of Prosody Encoder, the input speech $\mathbf{X_2}$ has to be any other speech sentence uttered by the same speaker. Similar to TTS, because the prosodies of input speech of Prosody Encoder and target converted speech do not match, the generated speech cannot be produced well. In Subsection~\ref{subsec:speechnet++}, we will address this issue.

\subsection{Adding one more module for TTS and VC: Prosody Predictor}
\label{subsec:speechnet++}

We point out the issues of TTS and VC in Subsubsections~\ref{subsubsec:tts} and ~\ref{subsubsec:vc}.
%Some approaches can be incorporated into SpeechNet to disentangle prosody and content embeddings, such as adversarial training~\citep{chen2020aipnet,chou2018multi} or information bottleneck~\citep{qian2020unsupervised}. They are worth investigating in the future work. 
Here we address them by proposing a modified version of SpeechNet by simply adding one additional module, Prosody Predictor, to generate the estimated prosody embeddings of target speech according to content and speaker embeddings
\begin{align}
    \mathbf{V'_p} = P_P(\mathbf{V_c}, \mathbf{v_s}).
    \label{eq:PP}
\end{align}
During training, there is one additional loss to make the estimated prosody $\mathbf{V'_p}$ as close as possible to the original target prosody $\mathbf{V_p}$ generated by Prosody Encoder:
\begin{align}
    \textnormal{(In TTS)\ \ } L_{prosody\_TTS} &= \parallel E_P(\mathbf{X}) - P_P(E_T(\mathbf{Y}, \mathbf{v'_s}), \mathbf{v'_s}) \parallel^2.
    \label{eq:loss_prosody_tts}
\end{align}
\begin{align}
    \textnormal{(In VC)\ \ } L_{prosody\_VC} &= \parallel E_P(\mathbf{X_{12}}) - P_P(E_C(\mathbf{X_1}), E_S(E_P(\mathbf{X_2}))) \parallel^2.
    \label{eq:loss_prosody_vc}
\end{align}
Then the Audio Decoder takes $\mathbf{V'_p}$ as input in TTS and VC.
The modified version of SpeechNet is shown in Figure~\ref{fig:speechnet++} in Appendix.

With Prosody Predictor, during inference, the generation of speech does not need to rely on the prosody from the speech input. 
The overall TTS process becomes:
\begin{align}
    \mathbf{X'} &= D_A(P_P(E_T(\mathbf{Y}, \mathbf{v'_s}), \mathbf{v'_s}), E_T(\mathbf{Y}, \mathbf{v'_s})).
    \label{eq:tts_mod}
\end{align}
And the overall VC process becomes:
\begin{align}
    \mathbf{X'_{12}} &= D_A(P_P(E_C(\mathbf{X_1}), E_S(E_P(\mathbf{X_2}))), E_C(\mathbf{X_1})).
    \label{eq:vc_mod}
\end{align}
For the experiments in Section~\ref{sec:exp}, we use SpeechNet with Prosody Predictor. We also present experiments with SpeechNet without Prosody Predictor in Appendix~\ref{sec:exp_ordinary_speechnet}.

\section{Experimental setup}
\label{sec:exp_setup}

This section introduces the model architecture, input/output formats of data, datasets, and evaluation metrics used in this paper.

\subsection{Model architecture}
\label{subsec:architecture}

Transformers~\citep{vaswani2017attention} have achieved state-of-the-art performances on many NLP tasks~\citep{devlin2019bert,wang2018glue}. More recently, many tasks in the speech domain started to use transformer-based models, such as ASR~\citep{gulati2020conformer,dong2018speech,karita2019comparative}, SE~\citep{kim2020t,fu2020boosting,nicolson2020masked}, SC~\citep{katta2020s,safari2020self,shi2020t}, TTS~\citep{ren2020fastspeech,li2019neural,zeng2020aligntts} and VC~\citep{huang2020voice,liu2020voice,lin2020fragmentvc}.

In this work, we adopt Conformer~\citep{gulati2020conformer} layers as the architectures for Prosody Encoder, Content Encoder, Audio Decoder and Prosody Predictor. The convolution blocks in Conformer make it empirically better than Transformer for speech data. 
Speaker Encoder is a self-attention pooling model~\citep{lin2017structured,zhu2018self}. 
For Unit Encoder, Duration Predictor and Length Regulator in Text Encoder, we use similar Transformer-based architectures as those in FastSpeech 2~\citep{ren2020fastspeech}. In the original FastSpeech 2, since it performs single-speaker TTS, no additional speaker embedding is required. In our SpeechNet, we concatenate the unit token vectors and speaker embedding vector as the input of Duration Predictor for the length regulation. Finally, the ordinary Transformer layers are adopted for S2S Decoder in Text Decoder, and CTC Decoder in Text Decoder is a single-layer fully-connected network. 
We insert 1D-convolution blocks right before Content Encoder for down-sampling. It is commonly used in ASR to obtain more compact content information~\citep{gulati2020conformer,dong2018speech,karita2019comparative}. Accordingly, we apply a similar down-sampler on content embedding vectors produced by Text Encoder. We also use 1D-convolution blocks right after Prosody Predictor and Content Decoder in Audio Decoder for up-sampling.
More implementation details about the hyperparameters, optimization and code are provided in Appendix~\ref{sec:detail} and released code.

\subsection{Datasets and evaluation metrics}
\label{subsec:data}

We use the commonly adopted dataset and evaluation metric for each task. 

For ASR, LibriSpeech~\citep{panayotov2015librispeech} is a corpus of read English speech from audiobooks of the LibriVox project. We perform ASR on the ``train-clean-100" set for training, which contains 100-hour speech uttered by 251 speakers, and on the ``test-clean" set for evaluation. For easier comparison in our experiments, we use greedy decoding and do not use beam-search decoding and additional language model rescoring during the inference. We measure the performance of ASR by the word error rate (WER).

For SE, Nonspeech~\citep{hu2010tandem} is a widely used noise dataset containing 100 types of noises. We choose LibriSpeech ~\citep{panayotov2015librispeech} as clean speech and randomly augment with noises from Nonspeech with various SNRs to create paired data for both training and testing. During the training stage, the ``train-clean-100" set is augmented with SNR $\in \{3, 6, 9\}$, while in the testing stage, the ``test-clean" set is augmented with $SNR \in \{-8, -6, -4, -2, 0, 2, 4, 6, 8\}$, providing a more severe condition to test whether the model can generalize. For evaluation, we report SiSDR, PESQ\citep{rix2001perceptual}, and STOI~\citep{taal2010short}. The first one measures the scale-invariant signal-to-distortion ratio, and the others align well with human's perspective. We select the checkpoint of the best SiSDR on validation set for testing.

For SC, VoxCeleb1~\citep{nagrani2017voxceleb} contains speech uttered by 1,251 celebrities extracted from videos uploaded to YouTube. We take 100 speakers from the official training and testing sets in this paper. The speaker classification accuracy is used as the evaluation metric of SC.

For TTS, LibriTTS~\citep{zen2019libritts} is a multi-speaker English corpus of read English speech from the audiobooks of the LibriVox project. Utterances with significant background noise are excluded in LibriTTS. Montreal Forced Aligner~\citep{mcauliffe2017montreal} is used to obtain the durations of phonemes. %``train-clean-100" is used as the training set. We have two testing sets. One only contains \textit{seen speakers} in the training set, and the other contains \textit{unseen speakers} that are not in the training set. For seen speakers, 
We randomly select 200 utterances from the ``train-clean-100" set for testing and the remaining ones for training.
%to make all the speakers in the testing set appear in the training set. For unseen speakers, we use ``dev-clean" set for testing. 
The MSE between the output and groundtruth mel-spectrograms serves as the evaluation metric of TTS. 

For VC, CMU Arctic~\citep{kominek2004cmu} is a corpus which consists of speech utterances by 18 speakers recorded under studio conditions. Every speaker utters the same set of sentences. In this paper, 1133 pairs of data are selected as testing data, and the remaining are for training.
%16 speakers are selected for training, and the other 1 male and 1 female speakers for testing. 
%To test the generalizability, we also randomly select paired data from 2 male and 2 female speakers from VCTK corpus~\citep{yamagishi2019cstr} as another testing set. 
Each training instance is a pair of speech utterances by different speakers. The 
MSE between the output and groundtruth mel-spectrograms serves as the evaluation metric of VC.

For further evaluation of TTS and VC, we apply a linear transformation to recover linear-scale spectrograms and the Griffin-Lim vocoder~\citep{griffin1984signal} to convert spectrograms back to wav files for listening. 

For all speech, we use 16kHz sample rate and extract the 80-dim mel-spectrogram features with 25 ms window size and 10 ms hop size. Then we add the first- and second-order derivatives and apply the cepstral mean and variance normalization (CMVN) commonly used for ASR and SC in previous works. 
%Particularly during the training of ASR, we apply frequency and time masking augmentation~\citep{park2019specaugment} on the speech inputs.
%The speech outputs for SE, TTS and VC are the 80-dim mel-spectrogram. 
%However for VC, since during validation, we need to listen to a couple of speech pairs to examine their conversion quality, using the Griffin-Lim vocoder is too time consuming. Instead, we train an additional fully-connected layer to convert the mel-spectrograms into linear spectrograms, which can be further converted into wav files by the faster inverted short-time Fourier transform. Therefore, the MSE loss is computed on the linear spectrograms during the training of VC, while the MSE losses are computed on the mel-spectrograms for SE and TTS.
The text input units for TTS are phonemes, and the output text units for ASR are subword units by the Byte Pair Encoding (BPE)~\citep{sennrich2016neural}, so the cross entropy loss is computed on the subword units.
More details about the datasets are provided in Appendix~\ref{sec:dataset_details}.

\section{Experiments}
\label{sec:exp}

We conduct single-task, two-task, and five-task learning experiments with the five tasks described in Subsection \ref{subsec:tasks}. Moreover, We experiment with two popular optimization strategies for MTL, ``AutoLoss" and ``PCGrad", which are described in Appendix~\ref{sec:optim}, for all of the two-task learning experiments. We also have the ablation study of two optimization strategies for five-task learning.

\subsection{Single-task and two-task learning results}
\label{subsec:exp_two_task}

The single-task and two-task experiment results are shown in Table~\ref{table:two_task}.
Each column shows the evaluation performance with a specific metric of a task. The first row is the name of the evaluation task, and the second row is the evaluation metric. 
The down-/up-arrow beside the evaluation metric means the better performance results in lower/higher numbers of that metric. 
The diagonal cells with pink shadow are the single-task results. 
The off-diagonal cells represent the results of joint training with a specific auxiliary task\footnote{The columns are for the evaluated tasks, while the rows for the auxiliary tasks}. 
The best number on each metric is highlighted with bold font. 
If the numbers of multi-task are better than the single-task ones, the numbers are underlined.

\begin{wrapfigure}{R}{0.35\textwidth}
    \centering
    \centerline{\includegraphics[width=0.3\textwidth]{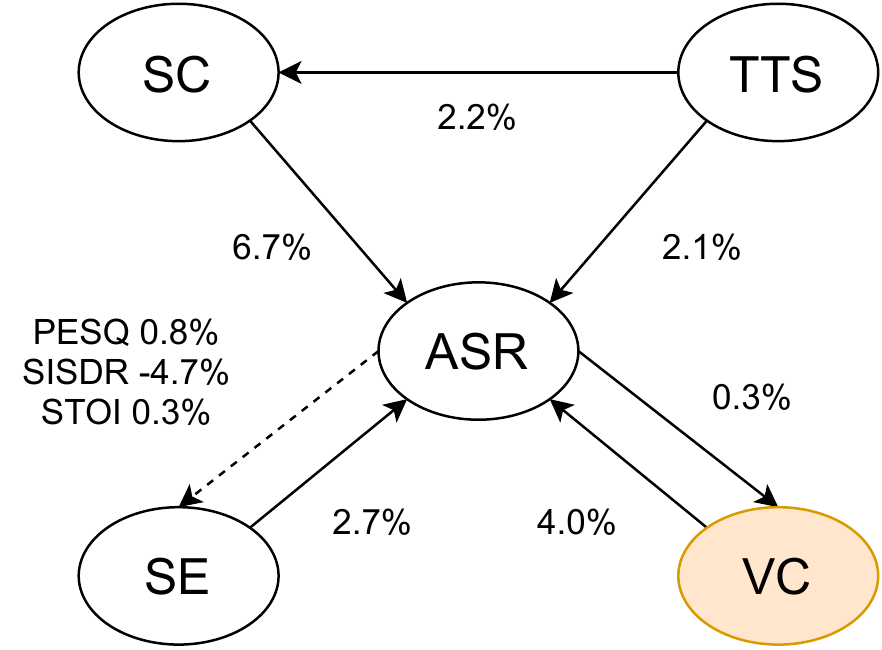}}
\caption{The improvement graph based on two-task learning results. We show the relative improvement beside the edge. The dashed line represents that PESQ and STOI are improved but SISDR is not.}
\label{fig:improvement}
\end{wrapfigure}

We also plot an "improvement graph" based on two-task learning results, as shown in Figure~\ref{fig:improvement}, to illustrate the beneficial relationship between all task pairs.
For example, if the model trained with ASR and SE improves the ASR WER compared to single-task ASR, we connect a directed edge from SE to ASR to denote the former benefits the latter.

\begin{table*}
\caption{\label{table:two_task} The results of single-task (pink cells) and two-task learning of five tasks.
}
\centering
\begin{tabular}{c|c|c|c|c|c|c|c}
\hline & \textbf{ASR} & \multicolumn{3}{c|}{\textbf{SE}} & \textbf{SC} & \multicolumn{1}{c|}{\textbf{TTS}} & \multicolumn{1}{c}{\textbf{VC}} \\ \hline
% & & \multicolumn{3}{c|}{} & & & \textbf{CMU} & \textbf{VCTK} \\ \hline
\textbf{Auxiliary} & \textbf{WER$\downarrow$} & \textbf{PESQ$\uparrow$} & \textbf{SISDR$\uparrow$} & \textbf{STOI$\uparrow$} & \textbf{ACC$\uparrow$} & \textbf{MSE$\downarrow$} & \textbf{MSE$\downarrow$} \\ \hline
ASR & \cellcolor{pink} 0.329 & \textbf{\underline{2.46}} & 5.62 & \textbf{\underline{0.880}} & 0.746 & 3.06 & \underline{5.93} \\
SE & \underline{0.320} & \cellcolor{pink} 2.44 & \cellcolor{pink} \textbf{5.90} & \cellcolor{pink} 0.877 & 0.820 & 3.08 & 6.06 \\
SC & \textbf{\underline{0.307}} & 2.15 & 4.02 & 0.850 & \cellcolor{pink} 0.860 & 2.98 & 6.04 \\
TTS & \underline{0.322} & 2.29 & 4.96 & 0.865 & \textbf{\underline{0.879}} & \cellcolor{pink} 2.94 & 6.02 \\
VC & \underline{0.316} & 2.02 & 4.80 & 0.847 & 0.703 & 3.57 & \cellcolor{pink} 5.95 \\
\hline
\end{tabular}
\end{table*}

We can observe ASR can be improved by all of the other tasks in two-task learning from the results. 
It indicates the related information provided by the other tasks can help Content Encoder to generate content embeddings for Text Decoder to produce text transcriptions. 
SE is improved slightly by ASR in terms of the PESQ and STOI metrics. 
The performance of SC is improved with the aid of TTS. 
VC is also slightly improved by ASR.

\subsection{Five-task learning results and ablation study of optimization strategies}
\label{subsec:exp_five_task}

\begin{table*}
\caption{\label{table:five_task} The ablation study of two optimization strategies with five-task learning of five tasks.
}
\centering
\begin{tabular}{c|c|c|c|c|c|c|c}
\hline & \textbf{ASR} & \multicolumn{3}{c|}{\textbf{SE}} & \textbf{SC} & \multicolumn{1}{c|}{\textbf{TTS}} & \multicolumn{1}{c}{\textbf{VC}} \\ \hline
% & & \multicolumn{3}{c|}{} & & & \textbf{CMU} & \textbf{VCTK} \\ \hline
\textbf{Optim Strategy} & \textbf{WER$\downarrow$} & \textbf{PESQ$\uparrow$} & \textbf{SISDR$\uparrow$} & \textbf{STOI$\uparrow$} & \textbf{ACC$\uparrow$} & \textbf{MSE$\downarrow$} & \textbf{MSE$\downarrow$} \\ \hline
AutoLoss\ +\ PCGrad & 0.511 & 1.99 & 3.71 & 0.837 & 0.451 & 3.36 & 6.01  \\
AutoLoss & 0.600 & 2.04 & 3.68 & 0.833 & 0.101 & 3.26 & \underline{5.88}  \\
PCGrad & 0.839 & 2.00 & 3.91 & 0.838 & 0.466 & 3.19 & 5.96  \\
No Strategy & 0.538 & 2.12 & 3.82 & 0.838 & 0.044 & 3.18 & \textbf{\underline{5.86}}  \\
\hline
\end{tabular}
\end{table*}

The ablation study of two optimization strategies with five-task learning is shown in Table~\ref{table:five_task}.
The optimization of five-task learning is much more difficult than single-task or two-task learning, as we can see, all of the performances degrade except for VC. 
Specifically, without PCGrad strategy, i.e. ``AutoLoss" and ``No Strategy" in the Table, SC cannot even be learned effectively. 
However, VC can be improved in these cases compared to single-task learning. 
With only ``PCGrad" strategy, ASR performs the worst. 
The model can learn all of the tasks only when using both AutoLoss and PCGrad strategies. 
But the results are still worse than single-task learning. 
This ablation study shows that optimization strategies influence the training of MTL significantly.
The two popular MTL optimization strategies in our experiments cannot help five-task learning outperform single-task learning.

%For subjective evaluation of TTS and VC, we randomly select a few wav file examples in the Supplementary Materials ``datasets\_and\_samples".

\subsection{Analysis and discussion}
\label{subsec:analysis}

In this paper, we exhibit five important speech processing tasks that can be learned with SpeechNet and conduct experiments of single-task, two-task, and five-task learning with two popular MTL optimization strategies.
However, many important research directions can be further extended from SpeechNet.
(1) As shown in the last subsection, suitable optimization strategies for MTL of speech processing tasks are important and desirable. SpeechNet can be a testbed for developing MTL optimization strategies on speech processing tasks.
(2) Besides MTL, other training schemes involving multiple tasks can be investigated in the future, such as transfer learning or meta learning. Different sizes of data are also worth investigating.
(3) SpeechNet is flexible and easy to modify or add new modules. Therefore many other speech or text processing tasks can be joined for research. For example, in our experiment results, ASR benefits from MTL the most. It may indicate other speech processing tasks that take speech input and produce text output can also benefit from MTL with SpeechNet, such as speech translation or multilingual speech recognition.

\begin{comment}
The first possible scheme is that multiple tasks generate augmentation data for one another.
In previous works for example, \citep{chen2015speech} uses SE to generate denoised features and helps ASR. \citep{tjandra2017listening,ren2019almost} trains both ASR and TTS, and the two tasks generate augmentation data for each other. \citep{hsu2019disentangling} uses SC to generate augmentation data for TTS. \citep{zhang2019improving} uses text supervision to generate augmentation data for VC.

Another possible direction is transfer learning.
For example, \citep{jia2018transfer} uses pretrained speaker verification model for TTS. \citep{huang2020voice,luong2019bootstrapping} pretrains TTS for VC.
Not only supervised tasks, self-supervised pretraining or semi-supervised learning~\citep{oord2018representation,baevski2020wav2vec,liu2020mockingjay} can be involved into the multi-task framework.
\end{comment}

\section{Conclusion}
\label{sec:conclusion}

In this paper, we propose a universal modularized model for speech processing tasks. We select five common and important tasks for multi-task learning experiments. 
The code and settings used in this paper are released to facilitate the research of modularized universal models or multi-task learning of speech processing tasks.

\begin{comment}
\begin{ack}
Use unnumbered first level headings for the acknowledgments. All acknowledgments
go at the end of the paper before the list of references. Moreover, you are required to declare
funding (financial activities supporting the submitted work) and competing interests (related financial activities outside the submitted work).
More information about this disclosure can be found at: \url{https://neurips.cc/Conferences/2021/PaperInformation/FundingDisclosure}.

Do {\bf not} include this section in the anonymized submission, only in the final paper. You can use the \texttt{ack} environment provided in the style file to autmoatically hide this section in the anonymized submission.
\end{ack}
\end{comment}

%\section*{References}

\bibliographystyle{plainnat}
%\bibliography{acl2021}

%%%%%%%%%%%%%%%%%%%%%%%%%%%%%%%%%%%%%%%%%%%%%%%%%%%%%%%%%%%%
\clearpage

\section*{Checklist}

\begin{enumerate}

\item For all authors...
\begin{enumerate}
  \item Do the main claims made in the abstract and introduction accurately reflect the paper's contributions and scope?
    \answerYes{}
  \item Did you describe the limitations of your work?
    \answerYes{} We point out the optimization difficulty of MTL in Section~\ref{sec:exp}.
  \item Did you discuss any potential negative societal impacts of your work?
    \answerNA{}
  \item Have you read the ethics review guidelines and ensured that your paper conforms to them?
    \answerYes{}
\end{enumerate}

\item If you are including theoretical results...
\begin{enumerate}
  \item Did you state the full set of assumptions of all theoretical results?
    \answerNA{}
	\item Did you include complete proofs of all theoretical results?
    \answerNA{}
\end{enumerate}

\item If you ran experiments...
\begin{enumerate}
  \item Did you include the code, data, and instructions needed to reproduce the main experimental results (either in the supplemental material or as a URL)?
    \answerYes{} Please see Section~\ref{sec:exp_setup}. Section~\ref{sec:exp}, Appendix and Supplementary Materials.
  \item Did you specify all the training details (e.g., data splits, hyperparameters, how they were chosen)?
    \answerYes{} Please see Section~\ref{sec:exp_setup}. Section~\ref{sec:exp}, Appendix and Supplementary Materials.
    \item Did you report error bars (e.g., with respect to the random seed after running experiments multiple times)?
    \answerNo{} 
	\item Did you include the total amount of compute and the type of resources used (e.g., type of GPUs, internal cluster, or cloud provider)?
    \answerYes{} Please see Appendix and Supplementary Materials.
\end{enumerate}

\item If you are using existing assets (e.g., code, data, models) or curating/releasing new assets...
\begin{enumerate}
  \item If your work uses existing assets, did you cite the creators?
    \answerYes{} Please see Subsection~\ref{subsec:data}
  \item Did you mention the license of the assets?
    \answerYes{} Please see Appendix and Supplementary Materials.
  \item Did you include any new assets either in the supplemental material or as a URL?
    \answerYes{} We release the code of SpeechNet. Please see Appendix and Supplementary Materials.
  \item Did you discuss whether and how consent was obtained from people whose data you're using/curating?
    \answerNA{}
  \item Did you discuss whether the data you are using/curating contains personally identifiable information or offensive content?
    \answerNA{}
\end{enumerate}

\item If you used crowdsourcing or conducted research with human subjects...
\begin{enumerate}
  \item Did you include the full text of instructions given to participants and screenshots, if applicable?
    \answerNA{}
  \item Did you describe any potential participant risks, with links to Institutional Review Board (IRB) approvals, if applicable?
    \answerNA{}
  \item Did you include the estimated hourly wage paid to participants and the total amount spent on participant compensation?
    \answerNA{}
\end{enumerate}

\end{enumerate}

%%%%%%%%%%%%%%%%%%%%%%%%%%%%%%%%%%%%%%%%%%%%%%%%%%%%%%%%%%%%

\newpage
\clearpage

\appendix

%\section{SpeechNet++}
%\label{sec:speechnet++}

\begin{figure*}[tb!]
  \centering
  \centerline{\includegraphics[width=0.7\linewidth]{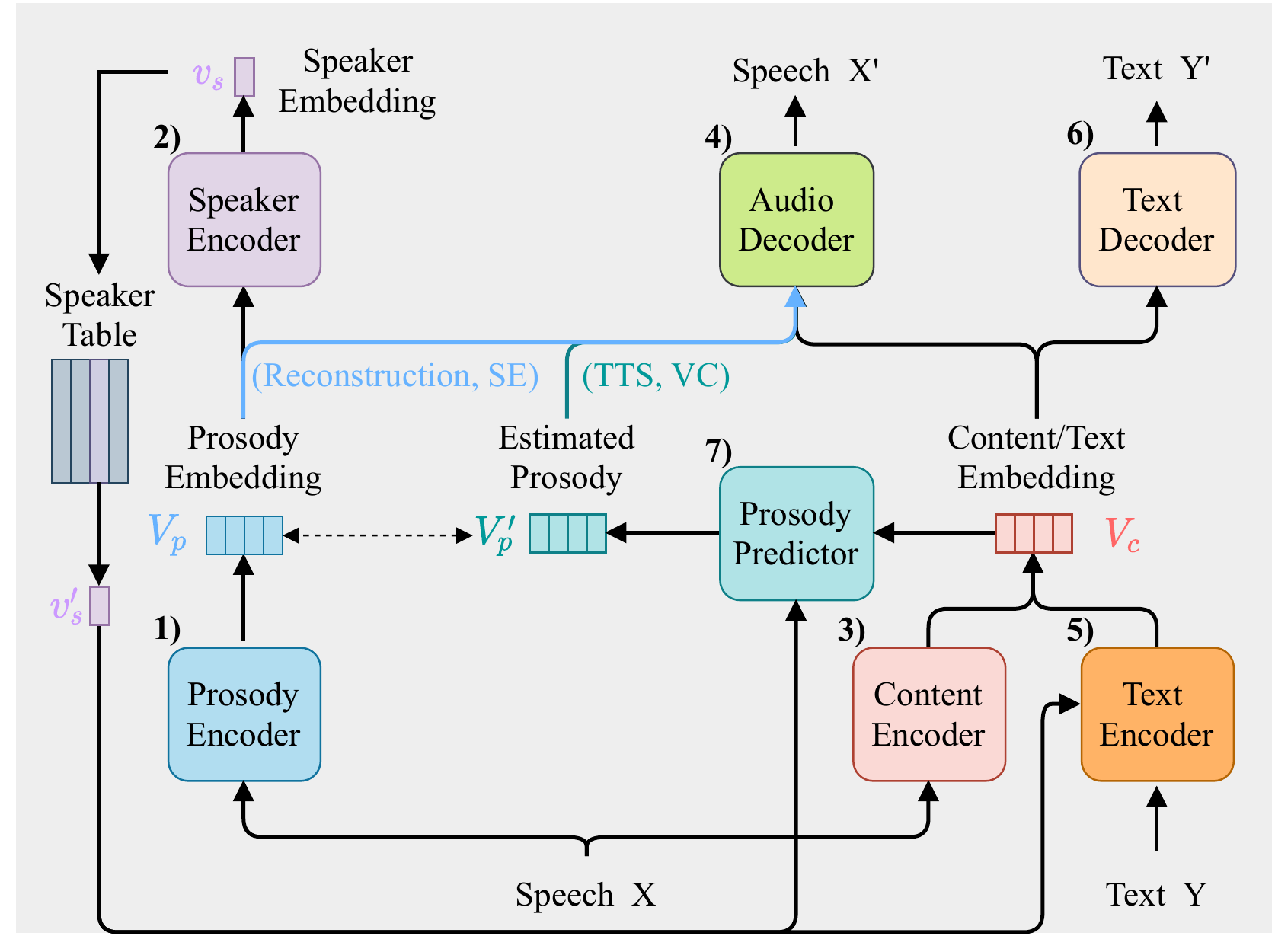}}
\caption{The modified architecture of SpeechNet by adding Prosody Predictor. }
\label{fig:speechnet++}
\end{figure*}

\section{Implementation details}
\label{sec:detail}

The batch size of each task is 16. We adopt AdamW optimizer~\citep{loshchilov2018fixing} with learning rate 3.0e-4, epsilon 1.0e-12 and betas [0.9, 0.999].
The dropout rate is set to 0.1 except for the dropout rate 0.5 in Duration Predictor in Text Encoder.
All the model parameters have decaying factor 0.01 except for those with names containing "bias", "norm-ff.weight", "norm-mha.weight", "norm-conv.weight" and "norm-final.weight".
The numbers of Conformer layers in each module are: Content Encoder 6, Speaker Encoder 3, Content Decoder in Audio Decoder 3, Merge Decoder in Audio Decoder 3, Unit Encoder in Text Encoder 4 and S2S Decoder in Text Decoder 4.
The CNN down- and up-samplers have 2 1d-convolution blocks with sample rate 4.
The $\alpha$ for ASR in Equation~\ref{eq:loss_asr} is 0.3.
The $\sigma$'s for the multi-task objective loss are initialized as 1.

The hidden dimension size is 256, and the linear unit size is 1024. The head number is 4 except for the head number 2 in Unit Encoder in Text Encoder. 
We use a linear learning rate warmup with 10000 steps and a linear decay with 100000 steps. For more details, please refer to the config file ``SpeechNet/config/libri/conformer\_256\_AdamW.yaml" in Supplementary Materials.

All the experiments are conducted on NVIDIA V100 GPUs. Each trial requires 2 GPUs with memory size 32GB.

\section{Details of datasets}
\label{sec:dataset_details}

The licenses of datasets used in this paper are stated as follows: LibriSpeech, VoxCeleb1 and LibriTTS are used under CC BY 4.0.
CMU Arctic is used under the license, ``Carnegie Mellon University, Copyright (c) 2003, All Rights Reserved. Permission to use, copy, modify, and license this software and its
 documentation for any purpose, is hereby granted without fee, subject to the following conditions: 1. The code must retain the above copyright notice, this list of conditions and the following disclaimer. 2. Any modifications must be clearly marked as such. 3. Original authors' names are not deleted. THE AUTHORS OF THIS WORK DISCLAIM ALL WARRANTIES WITH REGARD TO THIS SOFTWARE, INCLUDING ALL IMPLIED WARRANTIES OF MERCHANTABILITY
 AND FITNESS, IN NO EVENT SHALL THE AUTHORS BE LIABLE FOR ANY
 SPECIAL, INDIRECT OR CONSEQUENTIAL DAMAGES OR ANY DAMAGES WHATSOEVER RESULTING FROM LOSS OF USE, DATA OR PROFITS, WHETHER IN AN ACTION OF CONTRACT, NEGLIGENCE OR OTHER TORTIOUS ACTION, ARISING OUT OF OR IN CONNECTION WITH THE USE OR PERFORMANCE OF THIS SOFTWARE."

The 100 speakers in VoxCeleb1 selected for SC are from id10001 to id10099, provided in ``SpeechNet/corpus/VoxCeleb1-100/veri\_test\_class.txt" in Supplementary Materials.
The TTS testing set is provided in ``SpeechNet/corpus/LibriTTS-100/val.txt" in Supplementary Materials.
The testing sets in CMU Arctic are aew and slt.

\section{Optimization Strategies}
\label{sec:optim}

During MTL, some tasks may share some modules, and the gradients computed from different objective functions of these tasks are accumulated to update the shared modules. However, there are two problems: (1) how to balance these objective functions with different types and scales, and (2) how to deal with conflicting gradients of parameters between different tasks. We experiment with two popular MTL optimization strategies, tackling these two problems respectively.

\subsection{Loss balancing for MTL}
\label{subsec:loss}

 We adopt an automatic loss balancing technique (which we denote as ``AutoLoss" in the experiments) based on the task-dependent data-independent uncertainty measurement of each task~\citep{kendall2018multi}. It has been shown effective to capture the relative confidence between tasks and learn loss weights for tasks.

The overall objective function of $n$ objective functions can be defined as:
\begin{align}
    \Sigma_{i=1}^{n} L'_i = \Sigma_{i=1}^{n} (\frac{1}{\sigma_i^2}L_i + \log \sigma_i),
    \label{eq:scaled_loss}
\end{align}
where $L'_i$ is the scaled version of the original loss $L_i$ with the introduction of learnable scalar variables $\sigma$'s.
%The derivation and explanation of this formulation is at Appendix~\ref{sec:derivation}.

\subsection{Eliminating gradient conflicts in MTL}
\label{subsec:gradient}

PCGrad~\citep{yu2020gradient} is a gradient manipulation approach to handle conflicting gradients on the same set of parameters. Specifically, for parameters in every layer of the model, we perform PCGrad as illustrated in Algorithm \ref{alg:pcgrad}: If the gradients
between two tasks have negative cosine similarity, the gradient of one task is projected onto the normal plane of the gradient of the other task. In this way, the conflicting component of the gradient no longer exists.
\begin{algorithm}[H]
\label{alg:pcgrad}
\SetAlgoLined
\KwIn{Model parameters $\theta$, task minibatch $B = \{T_k\}$}
$\mathbf{g_k} \leftarrow \nabla_\theta L_k(\theta) \ \forall k$ \\
$\mathbf{g_k^{PC}} \leftarrow \mathbf{g_k} \ \forall k$ \\
 \For{$T_i \in B$}{
  \For{$T_j \sim B\textbackslash T_i$ in random order}{
   \If{$\mathbf{g_i^{PC}} \cdot \mathbf{g_j} < 0$}{
    Set $\mathbf{g_i^{PC}} = \mathbf{g_i^{PC}} - \frac{\mathbf{g_i^{PC}} \cdot \mathbf{g_j}}{\parallel \mathbf{g_j} \parallel^2} \mathbf{g_j}$
   }
  }
 }
 \KwOut{$\Delta \theta = \mathbf{g^{PC}} = \Sigma_i \mathbf{g_i^{PC}}$}
 \caption{PCGrad Update \citep{yu2020gradient}}
\end{algorithm}

\clearpage

\section{Experiments with SpeechNet without Prosody Predictor}
\label{sec:exp_ordinary_speechnet}

\begin{wraptable}{R}{0.4\textwidth}
\caption{\label{table:ordinary_speechnet} The results of single-task learning of TTS and VC with SpeechNet without Prosody Predictor.
}
\centering
\begin{tabular}{c|c}
\hline 
\multicolumn{1}{c|}{\textbf{TTS}} & \multicolumn{1}{c}{\textbf{VC}} \\ \hline
\textbf{MSE$\downarrow$} & \textbf{MSE$\downarrow$} \\ \hline
23.21 & 16.07 \\
\hline
\end{tabular}
\end{wraptable}

In this section, we show the single-task learning results of TTS and VC using SpeechNet without Prosody Predictor, i.e., using the training and testing procedures described in Subsubsections~\ref{subsubsec:tts} and~\ref{subsubsec:vc}. The results are shown in Table~\ref{table:ordinary_speechnet}. We can observe that the testing MSEs cannot be decreased because the prosodies of input speech of Prosody Encoder and target speech do not match. It validates our motivation and the necessity to generate the estimated prosody of target speech based on content and speaker embeddings.

\section{Randomly selected TTS and VC samples using SpeechNet with Prosody Predictor}
\label{sec:samples}

For subjective evaluation of TTS and VC using SpeechNet with Prosody Predictor, we randomly select a few genereated wav file examples in ``SpeechNet/TTS\_samples" and ``SpeechNet/VC\_samples" in Supplementary Materials.

The results in Table~\ref{table:two_task} and randomly selected samples in Supplementary Materials show that our proposed SpeechNet with Prosody Predictor can successfully learn TTS and VC.

\end{document}